\documentclass[sigconf,nonacm,screen]{acmart}

\setcopyright{none}
\acmDOI{}
\acmISBN{}

\AtBeginDocument{%
  }

\usepackage{booktabs}
\usepackage{multirow}
\usepackage{graphicx}
\usepackage{subcaption}
\usepackage{amsmath}
\usepackage{mathtools}
\usepackage{algorithm}
\usepackage{algpseudocode}
\usepackage{xcolor}
\usepackage{hyperref}
\usepackage{enumitem}

\newcommand{\doPair}{\textsc{DO} pair}
\newcommand{\CR}{\textsc{TC}}
\newcommand{\DI}{\textsc{DI}}

\newcommand{\DA}{\textsc{DA}}
\newcommand{\CBA}{\textsc{PCBA}}
\newcommand{\PCAA}{\textsc{PCAA}}
\newcommand{\DCAA}{\textsc{DCAA}}

\newcommand{\CAT}{\textsc{CAT}}

\newcommand{\hao}[1]{\PackageError{arxiv-draft}{Unresolved hao drafting note}{Resolve and remove the note from the source before publication.}}
\newcommand{\xu}[1]{\PackageError{arxiv-draft}{Unresolved xu drafting note}{Resolve and remove the note from the source before publication.}}

\definecolor{pos}{RGB}{0,120,60}
\definecolor{neg}{RGB}{180,40,40}
\definecolor{neu}{gray}{0.35}

\newcommand{\posdelta}[1]{\textcolor{pos}{#1}}
\newcommand{\negdelta}[1]{\textcolor{neg}{#1}}
\newcommand{\neudelta}[1]{\textcolor{neu}{#1}}

\newcommand{\methodname}{ALIGN-HOLD}

\begin{document}

\title[ALIGN-HOLD: Experience Alignment for Ride-Hailing Hold Control]{ALIGN-HOLD: Experience Alignment for Real-Time Hold Control in Large-Scale Ride-Hailing Matching at DiDi}

\author{Zuhao Zhang}
\authornote{Work done during an internship at Didichuxing Co. Ltd.}
\authornote{Equal contribution.}
\affiliation{%
  \institution{Shanghai Jiao Tong University}
  \city{Shanghai}
  \country{China}}
  \email{zhzhang5@sjtu.edu.cn}

\author{Xu Liu}
\authornotemark[2]
\affiliation{%
  \institution{Didichuxing Co. Ltd}
  \city{Beijing}
  \country{China}}
  \email{leoliuxu@didiglobal.com}

\author{Kai Wan}
\affiliation{%
  \institution{Didichuxing Co. Ltd}
  \city{Beijing}
  \country{China}}
  \email{peterwan@didiglobal.com}

\author{Zihao Lu}
\affiliation{%
  \institution{Didichuxing Co. Ltd}
  \city{Beijing}
  \country{China}}
  \email{luzihao@didiglobal.com}

\author{Li Ma}
\affiliation{%
  \institution{Didichuxing Co. Ltd}
  \city{Beijing}
  \country{China}}
  \email{malimarey@didiglobal.com}

\author{Shuai Li}
\affiliation{%
  \institution{Shanghai Jiao Tong University}
  \city{Shanghai}
  \country{China}}
  \email{shuaili8@sjtu.edu.cn}

\renewcommand{\shortauthors}{Zuhao Zhang et al.}

\begin{abstract}
Real-time \emph{hold control} is a high-leverage mechanism in large-scale ride-hailing systems: by selectively deferring driver--order pairs, the platform can wait for better matching opportunities and improve end-to-end passenger--driver experience.
Existing production systems such as EXHOLD learn bandit-based hold policies from handcrafted combinations of trip completion, cancellations, waiting time, and driver effort.
However, designing such rewards becomes increasingly difficult as marketplace preferences are heterogeneous and observed passenger--driver behavior can be sparse, noisy, and affected by dynamic supply--demand conditions.

We present \textbf{\methodname{}}, a production-scale experience alignment framework that learns hold policy from implicit marketplace preferences. \methodname{} constructs complementary preference pairs from order trajectories, driver trajectories, and contemporaneous local matching graphs, and trains an experience Reward Model (RM) using balanced multi-view sampling and model-adaptive hard preference sampling. During simulator-based policy learning, the frozen RM provides a dense, context-dependent reward and supports label-free filtering of low-identifiability interactions whose behavioral feedback is difficult to attribute to matching quality.

We deploy \methodname{} on DiDi's ride-hailing platform and evaluate it in a 28-day randomized A/B experiment, covering approximately $100{,}000$ passenger requests per day. Compared with the deployed production policy, \methodname{} achieves statistically significant improvements in trip completion rate and driver income, while significantly reducing passenger cancellations before and after driver acceptance. Complementary ablations, RM diagnostics, and behavioral analyses validate the contributions of the proposed components. \methodname{} has been fully ramped up and is currently serving DiDi's Brazil marketplace.
\end{abstract}



\maketitle

\section{Introduction}
\label{sec:intro}
\begin{figure*}[t]
    \centering
    \includegraphics[width=\linewidth]{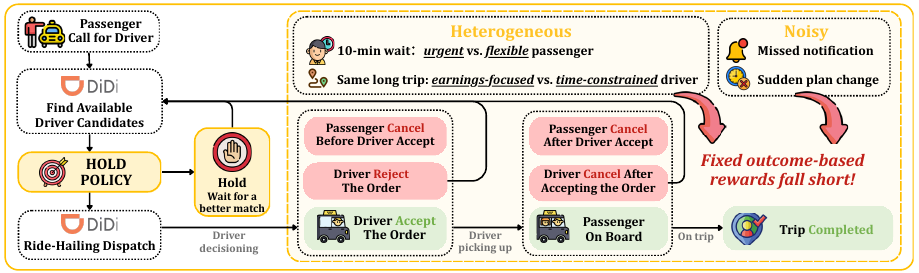}
    \vspace{-15pt}
    \caption{
Background and motivation for \textsc{\methodname{}}: learning hold control from heterogeneous, noisy feedback.
    }
    \label{fig:overview}
    \vspace{-10pt}
\end{figure*}

Large-scale ride-hailing platforms make real-time matching decisions in dynamic passenger--driver marketplaces~\cite{xu2018large}. As illustrated in Figure~\ref{fig:overview}, for each incoming order, the system evaluates candidate driver--order pairs (\doPair) and determines not only \emph{who} to match, but also \emph{when} each candidate should enter the downstream matching~\cite{wang2021secure,qin2021optimizing}. 
A critical mechanism in this pipeline is \emph{hold control}, which selectively defers certain \doPair{} candidates to await better matching opportunities.
An effective hold policy can reduce cancellations and excessive passenger waiting, mitigate wasted driver effort, and improve end-to-end trip success~\cite{suhr2019two}.
However, hold control is delicate: aggressive holding may suppress promising matches, whereas conservative holding yields little benefit.

Our previously deployed system, EXHOLD~\cite{liu2026exhold}, demonstrated that experience-aware hold control can deliver meaningful improvements in production.
EXHOLD first assigns each \doPair{} to an interpretable \emph{experience tier} with contextual bandits~\cite{chu2011contextual} and then converts the tier into an executable hold time through guardrail-constrained optimization. The bandit module is trained with a handcrafted reward that combines multiple experience-related outcomes, including trip completion, passenger and driver cancellations, waiting-related indicators, and driver effort.
This design provides a practical way to coordinate heterogeneous objectives and has been successfully deployed at scale~\cite{liu2025adaptive}.

However, the development of EXHOLD also reveals a fundamental bottleneck for further policy improvement: the experience objective remains manually specified.
While handcrafted rewards can approximate marketplace utility, they scale poorly across heterogeneous operating conditions. User preferences for factors such as waiting time, destination, and fare vary across regions and time periods~\cite{ashkrof2022ride,shi2023ride}.
A globally shared reward therefore requires frequent manual recalibration and may still fail to capture the marketplace’s latent experience objective~\cite{hadfield2017inverse}.
These limitations motivate us to shift from \emph{experience-aware reward engineering} to \emph{experience alignment}, where the optimization objective is learned directly from large-scale marketplace behaviors rather than manually predefined~\cite{christiano2017deep, liu2025online}.

However, learning such an alignment signal is challenging because ride-hailing systems provide only implicit and noisy preference feedback.
Unlike language-model alignment, where multiple responses to the same prompt can be generated and ranked by experts~\cite{ouyang2022training,zhang2026miniappbench, xia2024hallucination}, a ride-hailing platform observes only one realized interaction under each spatiotemporal state.
The feedback is thus sparse and weakly attributable: driver non-response dominates many broadcast sequences, explicit cancellation signals are relatively rare, and observed outcomes may be caused by external factors beyond the intrinsic compatibility of a \doPair{}.
Figure~\ref{fig:overview} illustrates such cases: a driver may fail to respond because of a missed notification, while a passenger may cancel because of an external change of plans. Neither behavior necessarily indicates an unfavorable match.
Therefore, directly treating behavioral outcomes as equally reliable supervision may cause the model to learn incidental noise rather than generalizable experience patterns~\cite{cheng2024rime}.

Despite lacking explicit preference labels, production trajectories contain rich \textit{relative} experience signals.
For example, an order may be canceled after broadcast to one driver but later matched to another driver and completed successfully. Although neither interaction provides an absolute satisfaction score, their relative outcomes in comparable contexts indicate which interaction better aligns with favorable passenger--driver experience.
Therefore, we construct preference pairs from such comparisons in order trajectories, driver trajectories, and local bipartite matching graphs, transforming isolated outcome labels into scalable implicit preference supervision for experience alignment~\cite{christiano2017deep}.

Building on this consideration, we propose \textbf{\textsc{\methodname{}}}, a deployable experience-alignment framework for real-time hold control. As illustrated in Figure~\ref{fig:method_pipeline}, \methodname{} constructs complementary preference pairs from three views---order-centric, driver-centric, and market-context.
It then trains an experience reward model (RM) on these pairs via model-adaptive hard preference sampling. The learned RM guides contextual-bandit policy learning in two complementary ways: by providing a dense, context-dependent alignment reward and filtering low-identifiability samples. The resulting policy outputs interpretable experience tiers, which are converted into executable hold times through the same production constrained-optimization layer as EXHOLD.

\paragraph{Real-world deployment and key findings.}
\methodname{} has been deployed in DiDi's production ride-hailing platform and evaluated against the production baseline through large-scale online A/B experiments across multiple Brazilian cities.
The results demonstrate that learning experience alignment from marketplace behaviors can further improve our hold-control system, leading to consistent gains in trip completion and matching efficiency while reducing experience-degrading cancellations.
Beyond aggregate improvements, we provide extensive ablations and behavioral analyses to understand how preference supervision, reward modeling, and policy alignment contribute to the observed gains.
Following successful online validation, \methodname{} has been fully ramped up and is now serving real traffic across the Brazilian marketplace.

\paragraph{Contributions.}

We summarize our main contributions as follows:
\begin{itemize}[leftmargin=*]
    \item We formulate the evolution of industrial hold control from handcrafted reward engineering to \emph{experience alignment}, addressing the challenge of learning policy objectives from sparse, heterogeneous, noisy, and confounded marketplace behavior.

    \item We propose a multi-view preference construction framework that derives scalable implicit pairwise preference supervision, together with model-adaptive hard preference sampling for robust reward model training.

    \item We develop a reward-model-guided policy alignment framework that uses the learned RM in two complementary ways: to provide a dense, context-dependent alignment reward and to dynamically filter low-identifiability samples during policy learning.

    \item We validate \methodname{} with large-scale online A/B experiments across multiple Brazilian cities. Experiment results demonstrate consistent improvements in trip success and matching efficiency while reducing experience-degrading cancellations. Ablations and behavioral analyses further verify the effectiveness of each component, and the system has been fully deployed in production.
\end{itemize}
\section{Problem Formulation}
\label{sec:problem}

We formulate real-time hold control as an experience-aligned decision problem in a dynamic ride-hailing marketplace. For each candidate \doPair{}, the platform determines whether the pair should immediately enter downstream matching or be temporarily held. The objective is to improve trip success and passenger--driver experience without unnecessarily holding promising matches.

\subsection{Real-Time Hold Decisions}
\label{sec:problem_hold}

Let $o_t\in\mathcal O$ denote a passenger order, $d_t\in\mathcal D$ an available driver, and $c_t$ the marketplace context at step $t$. The decision state is $s_t=(o_t,d_t,c_t)\in\mathcal S$,
where $c_t$ includes the information available at decision time, such as passenger and driver attributes, pickup and trip characteristics, and local supply--demand conditions. Given $s_t$, the hold policy assigns a discrete \emph{experience tier}:
\begin{equation}
    a_t\sim\pi_\theta(\cdot\mid s_t),
    \quad
    a_t\in\mathcal A=\{0,1,\ldots,K\}.
\end{equation}

Lower tiers indicate more favorable candidates for immediate exposure, whereas higher tiers indicate stronger preference for deferral. The selected tier is mapped to a hold duration $\tau_{a_t}$ by the guardrail-constrained execution layer inherited from EXHOLD~\cite{liu2026exhold}.

\subsection{Implicit Experience Feedback}
\label{sec:problem_preference}

Let $u^*(s,a)$ denote the latent passenger--driver utility of selecting tier $a$ under state $s$. Ideally, the policy would optimize
\begin{equation}
\label{eq:latent_policy_objective}
    \max_\theta\;
    \mathbb E_{
        s_t\sim\rho_{\pi_\theta},\,
        a_t\sim\pi_\theta(\cdot\mid s_t)
    }
    \left[
        u^*(s_t,a_t)
    \right],
\end{equation}
where $\rho_{\pi_\theta}$ is the state distribution induced by the evolving policy and marketplace. However, $u^*$ is not directly observable. The platform records only downstream behavioral outcomes, such as Trip Completion (\CR), driver non-response (non-\DA), Passenger Cancellation Before Driver Acceptance (\CBA), Passenger Cancellation After Driver Acceptance (\PCAA), and Driver Cancellation After Acceptance (\DCAA). 

EXHOLD approximates the latent utility using a handcrafted reward that manually combines behavioral outcomes and shaping terms. We instead learn a ranking surrogate $r_\phi(s,a)$ from implicit relative preferences, providing dense and context-dependent supervision for experience-aligned policy learning~\cite{biyik2018batch,christiano2017deep}. This formulation introduces two challenges: extracting reliable preferences from sparse and confounded marketplace behavior, and learning a robust policy from an imperfect learned reward.
\section{Method}
\label{sec:method}

\begin{figure*}[t]
    \centering
    \includegraphics[width=0.92\linewidth]{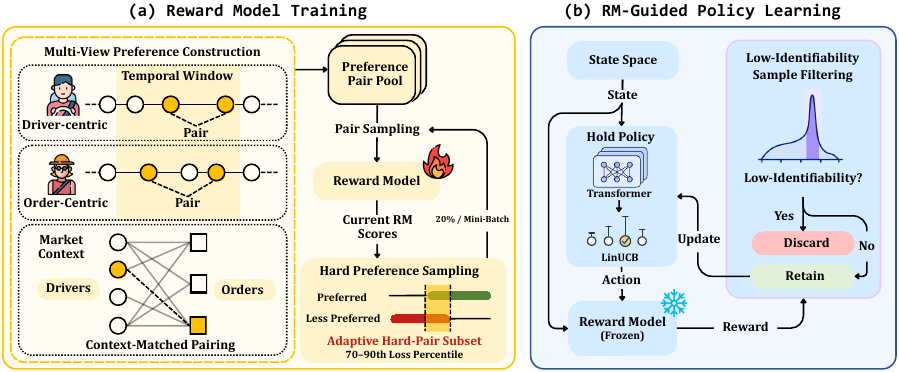}
    \caption{
    Training pipeline of \methodname{}.
    \textbf{(a)} Multi-view implicit preferences are used to train the Reward Model with model-adaptive hard preference sampling.
    \textbf{(b)} During interactive policy learning, the frozen RM scores state--action interactions, filters low-identifiability feedback, and provides dense rewards for updating the contextual-bandit policy.
    }
    \label{fig:method_pipeline}
\end{figure*}

We introduce \methodname{}, a deployable experience-alignment framework that learns an experience Reward Model (RM) from implicit marketplace preferences and uses it to guide contextual-bandit policy learning. As illustrated in Figure~\ref{fig:method_pipeline}, the framework contains two learning stages. First, we construct \textit{multi-view preference pairs}, and train the RM with balanced and \textit{model-adaptive preference sampling}. Second, the frozen RM provides dense rewards and \textit{filters low-identifiability feedback} as the evolving policy interacts with a matching simulator. The learned policy outputs discrete experience tiers, which are converted into hold times through the production execution layer shared with EXHOLD.
The complete training and deployment workflow is summarized in Appendix~\ref{sec:appendix_algorithm}.

\subsection{Multi-View Preference Construction}
\label{sec:preference_mining}

Ride-hailing platforms can not observe explicit preference labels because only one matching decision is realized under each marketplace state. Nevertheless, related interactions can provide comparative evidence when they share an order, a driver, or a local marketplace context. Therefore, we utilize these relationships to construct implicit preference pairs.

Let $(s_i,a_i,y_i)$ denote a logged interaction, where $s_i$ is the observed driver--order state, $a_i$ is the executed experience tier, and $y_i$ is the downstream outcome. We distinguish trip completion, denoted by $\CR$, from the non-completion outcomes
\begin{equation}
    \mathcal Y^{-}
    =
    \{
        \text{non-}\DA,
        \CBA,
        \PCAA,
        \DCAA,
        \text{other non-}\CR
    \},
    \label{eq:outcome_partition}
\end{equation}
and apply the same preference rule across all views:
\begin{equation}
\label{eq:preference_rule}
    (s_i,a_i)\succ(s_j,a_j)
    \quad \text{if} \quad
    y_i=\text{\CR}
    \ \ \text{and} \ \ 
    y_j\in\mathcal Y^{-}.
\end{equation}
Trip completion provides the strongest observable signal of joint interaction success. We therefore treat a completed decision as preferred to a related non-completed decision, while avoiding specified ordering among different failure outcomes, as doing so would reintroduce the manually specified trade-offs that the learned reward is intended to avoid. Under this shared completion-based ordering, we construct preference pairs from three complementary views, each reducing a different source of contextual variation.

\paragraph{Order-centric view.}
We compare completed and non-completed interactions from the same order trajectory. For example, an unsuccessful broadcast followed by a later interaction that completes the trip forms a preference pair. Sharing the order reduces variation in request-level factors such as origin, destination, and passenger intent, while retaining differences across candidate drivers, experience-tier actions, and evolving matching states.

\paragraph{Driver-centric view.}
We compare completed and non-completed interactions involving the same driver within a bounded temporal window. Sharing the driver reduces variation in persistent driver characteristics, while the temporal restriction limits changes in availability and working conditions. The resulting pairs provide relative evidence across candidate orders and local contexts faced by the same driver.

\paragraph{Market-context view.}
We compare completed and non-completed interactions from the same or temporally adjacent local matching-graph snapshots. A snapshot contains active orders, available drivers, and their eligible candidate edges within a dispatch round. These comparisons provide relative evidence among contemporaneous alternatives under similar local supply--demand conditions.

For each view $v\in\{\mathrm{ord},\mathrm{drv},\mathrm{mkt}\}$, we obtain a preference pool
\begin{equation}
    \mathcal P_v
    =
    \left\{
        \left(
            (s_i^+,a_i^+),(s_i^-,a_i^-)
        \right)
    \right\}_{i=1}^{N_v},
\end{equation}
where each element pairs a preferred completed decision $(s_i^+,a_i^+)$ with a related non-completed decision $(s_i^-,a_i^-)$. The three pools provide complementary preference evidence from the order, driver, and marketplace perspectives, while Equation~\eqref{eq:preference_rule} supplies a shared rule for assigning their preference labels. Since the pools can differ substantially in size, we balance their contributions during RM training through view-level sampling at a $1{:}1{:}1$ ratio, as described in the next subsection. Detailed trajectory windows and pairing criteria are provided in Appendix~\ref{app:preference_rules}.

\subsection{Experience Reward Model}
\label{sec:reward_model}

The RM $r_\phi(s,a)$ assigns a scalar experience score to each state--action decision. We parameterize it as an MLP--Transformer scorer: an input MLP embeds the state--action features into latent tokens, a Transformer encoder models cross-feature interactions, and an MLP head produces the final score.
Detailed architecture and training configurations are provided in Appendix~\ref{sec:appendix_rm_settings}.

Given a preferred decision $(s^+,a^+)$ and a less preferred decision $(s^-,a^-)$, we adopt the Bradley--Terry preference model~\cite{bradley1952rank}:
\begin{equation}
    P_\phi\left(
        (s^+,a^+)\succ(s^-,a^-)
    \right)
    =
    \sigma\left(
        r_\phi(s^+,a^+)-r_\phi(s^-,a^-)
    \right),
\end{equation}
where $\sigma(\cdot)$ is the sigmoid function. The pairwise loss is
\begin{equation}
\label{eq:rm_loss}
    \ell_\phi
    =
    -
    \log
    \sigma\left(
        r_\phi(s^+,a^+)-r_\phi(s^-,a^-)
    \right).
\end{equation}

Since this objective constrains relative score differences rather than an absolute utility scale, $r_\phi$ is interpreted as an experience-ranking signal rather than a calibrated satisfaction value.

\paragraph{Balanced and model-adaptive sampling.}
The three preference pools in Section~\ref{sec:preference_mining} differ substantially in size; sampling directly from their union would allow the largest view to dominate RM training. We therefore allocate $80\%$ of each mini-batch to regular samples drawn from the order-centric, driver-centric, and market-context pools at a $1{:}1{:}1$ ratio. The remaining $20\%$ are selected through model-adaptive hard preference sampling: we measure pair difficulty using the current loss in Equation~\eqref{eq:rm_loss} and sample globally from pairs between the $70$th and $90$th percentiles of the mixed-pool loss distribution. This interval emphasizes hard but informative comparisons while avoiding the extreme high-loss tail, which is more likely to contain incorrectly inferred or heavily confounded preferences~\cite{katharopoulos2018not,jiang2018mentornet,cheng2024rime}. Since pairwise losses are recomputed as the RM evolves, the hard-pair distribution adapts throughout training. The contribution of balanced multi-view and hard preference sampling is evaluated in Section~\ref{sec:ablation}.

\subsection{RM-Guided Policy Alignment}
\label{sec:policy_alignment}

After RM training, we freeze $r_\phi$ and use it both to provide a dense learning reward and to remove feedback for which the observed state--action representation offers weak outcome evidence.

\paragraph{Dense alignment reward.}
We normalize RM scores using statistics from a held-out calibration set:
\begin{equation}
\label{eq:normalized_reward}
    \widetilde r_\phi(s,a)
    =
    \frac{
        r_\phi(s,a)-\mu_\phi
    }{
        \sigma_\phi+\epsilon
    },
\end{equation}
where $\mu_\phi$ and $\sigma_\phi$ are the calibration-set mean and standard deviation. Normalization preserves the learned ordering while providing a stable reward scale. Unlike the binary labels used to construct preference pairs, $\widetilde r_\phi(s,a)$ provides a continuous, context-dependent signal: it allows the policy to generalize relative completion evidence across driver--order states and experience-tier actions rather than directly optimizing a sparse completion indicator.

\paragraph{RM-guided low-identifiability filtering.}
Our starting intuition is that not all behavioral feedback can be reliably attributed to the quality of a matching decision. In particular, under homogeneous or low-cost matching contexts, different realized outcomes may arise from unobserved user intent or incidental behavior, making the underlying experience preference difficult to infer from state--action trajectories. We observe, however, that they tend to form a dominant high-density band in the frozen RM-score distribution, where the RM assigns similar scores to interactions with otherwise heterogeneous outcomes. We therefore use this density concentration as a label-free proxy for weakly attributable feedback. For each policy-training slice $b$, we dynamically detect the corresponding score band $\mathcal B_{\mathrm{low}}^{(b)}$ and exclude interactions within it from policy updates. Section~\ref{sec:rm_analysis} validates on held-out data that the detected bands exhibit strong cross-outcome overlap and coherent low-identifiability feature profiles. Implementation details about the detection of the low-identifiability region are provided in Appendix~\ref{sec:appendix_policy_settings}.

\paragraph{Interactive policy learning.}
With the RM frozen, the hold policy is trained through continuous interaction with the same production-calibrated matching simulator used by EXHOLD~\cite{liu2026exhold}. The simulator models real-time candidate generation and state transitions using historical marketplace dynamics; its construction is inherited from the existing production pipeline and is not a contribution of this work. At step $t$ within training slice $b$, the simulator provides state $s_t$, the current policy selects an experience tier $a_t=\pi_\theta(s_t)$, and the RM produces the raw score $u_t=r_\phi(s_t,a_t)$ and normalized reward $\widetilde r_\phi(s_t,a_t)$. We retain the interaction according to
\begin{equation}
\label{eq:retention}
    m_t
    =
    \mathbb I
    \left\{
        u_t \notin \mathcal B_{\mathrm{low}}^{(b)}
    \right\}.
\end{equation}
Interactions with $m_t=1$ provide $\widetilde r_\phi(s_t,a_t)$ as the contextual-bandit reward, whereas those with $m_t=0$ are excluded from policy updates. As the policy evolves, newly generated interactions follow its changing state--action distribution, allowing RM-guided supervision to adapt throughout training. Appendix~\ref{app:simulator_fidelity} provides further discussion of simulator evaluation and offline--online fidelity.

\subsection{Production Decision and Execution}
\label{sec:execution}

The deployed \methodname{} system retains the production decision and execution architecture of EXHOLD~\cite{liu2026exhold}. Its online path consists of two decoupled components: (i) a Transformer--LinUCB~\cite{chu2011contextual,vaswani2017attention} policy $\pi_\theta$ that maps each driver--order state $s_t$ to an experience tier $a_t\in\{0,\ldots,K\}$, and (ii) a tier-to-time table produced by the existing guardrail-constrained optimizer. The table satisfies
\begin{equation}
    0\leq\tau_0\leq\tau_1
    \leq\cdots\leq\tau_K
    \leq\tau_{\max},
\end{equation}
where $\tau_k$ is the hold duration assigned to tier $k$; details are organized in Appendix~\ref{sec:appendix_hold_calibration}. At serving time, the system performs $a_t=\pi_\theta(s_t), \tau_t=\tau_{a_t}$,
and passes $\tau_t$ to the downstream matching pipeline.
\section{Experiments}

\begin{table*}[t]
    \centering
    \caption{
    Online A/B results of \methodname{} versus the deployed EXHOLD baseline.
    We report relative treatment--control deltas (T$-$C).
    Positive values are desirable for metrics marked with $\uparrow$, whereas negative values are desirable for metrics marked with $\downarrow$.
    $^*$ indicates statistical significance at $p<0.05$.
    }
    \label{tab:main_online_results}
    \begin{tabular}{lccc}
        \toprule
        \textbf{Metric} & \textbf{Delta (T$-$C)} & \textbf{$p$-value} & \textbf{Significance} \\
        \midrule
        \multicolumn{4}{l}{\textbf{Core Marketplace Outcomes}} \\
        Trip Completion (\CR) ratio $\uparrow$
        & \textbf{\posdelta{+0.57\%}} & 0.00011 & $^*$ \\
        Driver Income (\DI) $\uparrow$
        & \textbf{\posdelta{+0.64\%}} & 0.00906 & $^*$ \\
        \midrule
        \multicolumn{4}{l}{\textbf{Passenger--Driver Experience}} \\
        Driver Acceptance (\DA) $\uparrow$
        & \textbf{\posdelta{+0.51\%}} & 0.00260 & $^*$ \\
        Passenger Cancellation Before Acceptance (\CBA) $\downarrow$
        & \textbf{\posdelta{-1.85\%}} & 0.00172 & $^*$ \\
        Passenger Cancellation After Acceptance (\PCAA) $\downarrow$
        & \textbf{\posdelta{-2.07\%}} & 0.00598 & $^*$ \\
        Driver Cancellation After Acceptance (\DCAA) $\downarrow$
        & \posdelta{-1.35\%} & 0.05137 &  \\
        \midrule
        \multicolumn{4}{l}{\textbf{Matching Efficiency and Service}} \\
        Call-Acceptance Time (\CAT) $\downarrow$
        & \posdelta{-0.20\%} & 0.09164 &  \\
        Accepted / Broadcasted $\uparrow$
        & \textbf{\posdelta{+1.41\%}} & 0.00117 & $^*$ \\
        Accepted / Called $\uparrow$
        & \textbf{\posdelta{+1.26\%}} & 0.00069 & $^*$ \\
        \midrule
        \multicolumn{4}{l}{\textbf{Operational Guardrail}} \\
        Overall Hold Ratio
        & \neudelta{+0.04\%} & -- & -- \\
        \bottomrule
    \end{tabular}
\end{table*}

\label{sec:experiments}

We evaluate \methodname{} through large-scale production experiments and complementary analyses. Our experiments are designed to answer four questions:
\textbf{(RQ1)} Does \methodname{} improve marketplace outcomes and passenger--driver experience over the deployed EXHOLD policy?
\textbf{(RQ2)} Which components contribute to its online gains?
\textbf{(RQ3)} Does the RM provide a meaningful alignment signal, and what feedback is removed by low-identifiability filtering?
\textbf{(RQ4)} How does the aligned policy reallocate hold decisions across driver--order characteristics and marketplace conditions?

\subsection{Experimental Setup}
\label{sec:setup}

\paragraph{Online A/B experiment.}
We compare \methodname{} with the deployed EXHOLD policy in DiDi's production ride-hailing system. EXHOLD learns its experience-tier policy from a handcrafted multi-objective reward, whereas \methodname{} replaces this supervision with the proposed RM-guided alignment pipeline. Both systems use the same policy architecture, matching simulator, and guardrail-constrained tier-to-time execution layer. The comparison therefore isolates the effect of experience-aligned policy learning.

The randomized A/B experiment runs from May~31 to June~27, 2026, covering 28 days and five cities in Brazil. Eligible traffic is assigned to treatment and control using the production 30-min time-slice protocol. The experiment covers approximately $100{,}000$ passenger requests per day. We apply two-sided statistical tests and regard $p<0.05$ as statistically significant.

\paragraph{Evaluation metrics.}
Our primary marketplace outcomes are Trip Completion (\CR) and Driver Income (\DI). We measure two-sided experience using Driver Acceptance (\DA), Passenger Cancellation Before Acceptance (\CBA), Passenger Cancellation After Acceptance (\PCAA), and Driver Cancellation After Acceptance (\DCAA). Call-Acceptance Time (\CAT), acceptance-funnel conversion, and the overall hold ratio measure operational safety.

\paragraph{Offline RM evaluation.}
Preference data are constructed from multi-view preference in Section~\ref{sec:preference_mining}. The data are chronologically divided into training, validation, and test sets using a $7{:}2{:}1$ ratio, with trajectory-level entities restricted to a single split to reduce preference leakage. Because no ground-truth scalar experience label exists, we evaluate the RM using held-out pairwise ranking accuracy and auxiliary outcome-discrimination metrics.

\subsection{RQ1: Overall Online Evaluation}
\label{sec:online_results}

Table~\ref{tab:main_online_results} shows that \methodname{} significantly improves both core marketplace outcomes over the production-validated EXHOLD baseline: \CR{} increases by $0.57\%$, while \DI{} increases by $0.64\%$. Since the two systems share the same policy architecture and hold-time execution layer, these improvements can be primarily attributed to the learned alignment signal and the resulting policy learning.

The funnel metrics provide a consistent account of the completion gain. \DA{} increases by $0.51\%$, while Accepted/Broadcasted and Accepted/Called improve by $1.41\%$ and $1.26\%$, respectively. Passenger-side cancellations also decrease substantially: \CBA{} falls by $1.85\%$ and \PCAA{} by $2.07\%$. \DCAA{} and \CAT{} improve directionally, although their changes do not reach statistical significance. Together, these results indicate that \methodname{} improves conversion and match stability across multiple stages rather than shifting cost from passengers to drivers. Importantly, the overall hold ratio changes by only $0.04\%$. The gains therefore do not result from globally increasing intervention intensity. Instead, the improved acceptance conversion and lower cancellations are consistent with a more precise allocation of hold decisions.

\paragraph{Cross-market consistency and operating regimes.}
We further examine whether the aggregate treatment effect is robust across heterogeneous production environments. Table~\ref{tab:segment_results} reports relative treatment--control deltas for the five experimental cities and for peak versus off-peak periods.

\begin{table}[t]
    \centering
    \caption{
    Performance of \methodname{} across cities and traffic periods. We report treatment--control deltas (T $-$ C) for representative marketplace and experience metrics.
    }
    \label{tab:segment_results}
    \begin{tabular}{lrrrr}
        \toprule
        \textbf{Segment} &
        \textbf{\CR{} $\uparrow$} &
        \textbf{\DI{} $\uparrow$} &
        \textbf{\PCAA{} $\downarrow$} &
        \textbf{\CAT{} $\downarrow$} \\
        \midrule
        Overall
        & +0.57\%
        & +0.64\%
        & -2.07\%
        & -0.20\% \\
        \midrule
        City A
        & +0.42\%
        & +0.46\%
        & -1.81\%
        & -0.18\% \\
        City B
        & +0.55\%
        & +0.67\%
        & -1.89\%
        & -0.24\% \\
        City C
        & +0.83\%
        & +0.98\%
        & -2.78\%
        & -0.27\% \\
        City D
        & +0.59\%
        & +0.71\%
        & -1.92\%
        & -0.26\% \\
        City E
        & +0.47\%
        & +0.55\%
        & -1.83\%
        & -0.12\% \\
        \midrule
        Peak hours
        & +0.78\%
        & +0.85\%
        & -2.76\%
        & -0.38\% \\
        Off-peak hours
        & +0.43\%
        & +0.44\%
        & -1.81\%
        & -0.14\% \\
        \bottomrule
    \end{tabular}
\end{table}

As shown in Table~\ref{tab:segment_results}, the effects of \methodname{} are directionally consistent across all five cities: every market exhibits improved \CR{} and \DI{}, together with lower \PCAA{} and \CAT{}. Although the magnitudes vary across cities, no individual market determines the direction of the aggregate result. The temporal split in Table~\ref{tab:segment_results} further shows that \methodname{} achieves larger gains during peak hours across all four metrics. This pattern is consistent with our experience that hold decisions often has greater downstream leverage when traffic intensity is high and competition for supply makes poorly aligned exposure more costly. The policy nevertheless remains directionally beneficial during off-peak periods, indicating that its effectiveness is not restricted to dense traffic conditions.

\subsection{RQ2: Ablation Study}
\label{sec:ablation}

In this section, we evaluate the three main design choices that distinguish \methodname{}: multi-view preference construction, model-adaptive hard sampling, and low-identifiability region filtering. 

\subsubsection{End-to-End Policy Ablations}

Table~\ref{tab:online_ablation} shows that the three components play complementary roles. Using only order-centric preferences preserves modest gains, but performs substantially worse than the full method, indicating that driver-centric and market-context comparisons provide experience evidence not captured by order trajectories alone. Removing model-adaptive hard sampling further reduces \CR{} and reverses the gain in \DI{}, showing that difficult preference pairs are particularly important for learning driver-relevant distinctions. Retaining low-identifiability feedback produces a similar imbalance: completion improves slightly, but driver income decreases, suggesting that dense RM rewards alone remain vulnerable to ambiguous behavioral supervision. Overall, their combination yields the strongest balanced marketplace gains.

\subsubsection{Preference-View Ablations}

\begin{table}[t]
  \centering
  \caption{
  Online ablation of \methodname's main design choices. We report treatment--control deltas for TC and DI.
  }
  \label{tab:online_ablation}
  \begin{tabular}{lcc}
    \toprule
    \textbf{Variant} & \textbf{TC $\uparrow$} & \textbf{DI $\uparrow$} \\
    \midrule
    \multicolumn{3}{l}{\textbf{RM Training Ablations}} \\
    w/o multi-view (order-centric only) 
    & \posdelta{+0.25\%}
    & \posdelta{+0.16\%} \\
    w/o model-adaptive hard sampling
    & \posdelta{+0.10\%}
    & \negdelta{-0.26\%} \\
    \midrule
    \multicolumn{3}{l}{\textbf{Policy Alignment Ablation}} \\
    w/o low-identifiability region filtering
    & \posdelta{+0.17\%}
    & \negdelta{-0.37\%} \\
    \midrule
    \methodname{} (full)
    & \textbf{\posdelta{+0.57\%}}
    & \textbf{\posdelta{+0.64\%}} \\
    \bottomrule
  \end{tabular}
\end{table}

Here we use two complementary diagnostics to evaluate the RM. \emph{Pairwise accuracy} measures the fraction of held-out preference pairs for which the RM assigns a higher score to the preferred DO. Besides, we report \emph{Macro Outcome AUC} on individual held-out interactions. For each non-completion category $y\in\mathcal Y^{-}$, we compute the binary AUC for distinguishing completed interactions from interactions with outcome $y$, and then average across outcome categories:
\begin{equation}
    \operatorname{OutcomeAUC}
    =
    \frac{1}{|\mathcal Y^{-}|}
    \sum_{y\in\mathcal Y^{-}}
    \operatorname{AUC}
    \left(
        \CR \ \mathrm{vs.}\ y
    \right).
\end{equation}

\begin{table}[t]
    \centering
    \caption{
    Multi-view preference ablations of \methodname{}.
    }
    \label{tab:view_ablation}
    \begin{tabular}{lcc}
        \toprule
        \textbf{Preference Data} &
        \textbf{Pair Acc. $\uparrow$} &
        \textbf{Macro AUC $\uparrow$} \\
        \midrule
        Order-centric only
        & 0.792 & 0.811 \\
        Driver-centric only
        & 0.774 & 0.803 \\
        Market-context only
        & 0.751 & 0.784 \\
        \midrule
        w/o order-centric
        & 0.806 & 0.819 \\
        w/o driver-centric
        & 0.814 & 0.825 \\
        w/o market-context
        & 0.820 & 0.833 \\
        \midrule
        All views
        & \textbf{0.845} & \textbf{0.852} \\
        \bottomrule
    \end{tabular}
\end{table}

Table~\ref{tab:view_ablation} shows that the three views provide complementary preference supervision. Among the single-view variants, the order-centric view performs best, consistent with comparisons within the same request reducing variation in trip intent and request context. Other single views are individually weaker but provide additional evidence about heterogeneous driver responses and contemporaneous matching alternatives. Besides, removing any view degrades the performance, whereas the balanced combination of all three views achieves the strongest performance.
Additional outcome-wise AUC results and RM-score-decile analyses are reported in Appendix~\ref{sec:rm_validation}.

\begin{figure*}[t]
    \centering
    \begin{subfigure}[t]{0.45\textwidth}
        \centering
        \includegraphics[width=\linewidth]{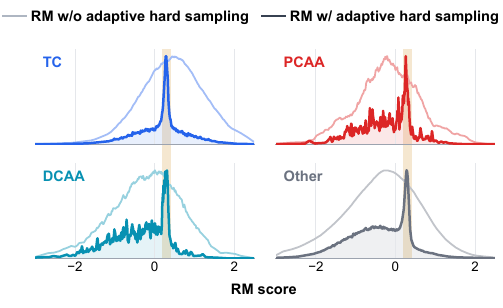}
        \caption{RM-score densities by downstream outcome.}
        \label{fig:rm_score_overlap}
    \end{subfigure}
    \hfill
    \begin{subfigure}[t]{0.52\textwidth}
        \centering
        \includegraphics[width=0.9\linewidth]{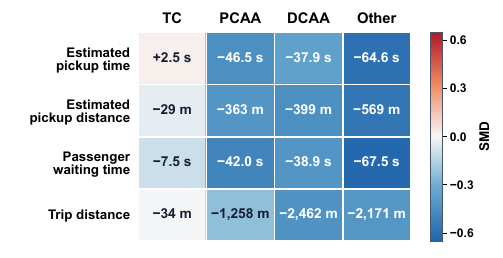}
        \caption{Feature shifts of low-identifiability samples.}
        \label{fig:filtered_features}
    \end{subfigure}
    \caption{
    Diagnostics of low-identifiability feedback.
    \textbf{(a)} Unit-peak RM-score densities by downstream outcome for RMs trained with uniform and model-adaptive hard preference sampling.
    \textbf{(b)} Within-outcome differences between interactions inside and outside the dynamically detected score bands; colors denote standardized differences and annotations report differences in the original units.
    }
    \label{fig:rm_diagnostics}
\end{figure*}

\subsection{RQ3: Alignment-Signal Reliability and Low-Identifiability Filtering}
\label{sec:rm_analysis}

In this section, we evaluate whether the density-based filtering used during simulator policy learning identifies interactions with weakly attributable behavioral feedback. For each simulator-training slice, the rejection band is estimated without outcome labels from the training partition and then examined on held-out interactions. Outcome categories are used only for this diagnostic analysis; they neither determine the detected band nor enter policy updates.

Figure~\ref{fig:rm_diagnostics}(a) compares held-out score distributions from RMs trained with uniform and model-adaptive hard preference sampling. Hard preference sampling improves learning from difficult comparisons and also makes a localized density mode shared by \CR{}, \PCAA{}, \DCAA{}, and other non-completion outcomes more apparent. Because distinct outcomes concentrate around similar scores, interactions near this mode provide limited evidence for distinguishing successful from unsuccessful matching decisions. This observation supports detecting the dominant score-density spike independently within each simulator slice, rather than using an outcome-specific or globally fixed threshold.

Figure~\ref{fig:rm_diagnostics}(b) further shows that the detected interactions form a coherent marketplace cohort. Within the major non-completion categories, interactions inside the detected bands have substantially shorter pickup time, passenger waiting time, pickup distance, and trip distance than retained interactions. The corresponding shifts within completed trips are comparatively small. Thus, low-identifiability feedback is concentrated among non-completed interactions whose observable waiting and travel costs are already low. In these contexts, the realized outcome may be more sensitive to unobserved intent or incidental behavior than to observable driver--order compatibility. These interactions are therefore not necessarily low quality, but their outcomes provide ambiguous supervision for the simulator-trained policy. Together, the score and feature diagnostics support excluding them from policy updates through the per-slice, label-free detector described in Appendix~\ref{sec:appendix_policy_settings}.

\begin{figure*}[t]
    \centering
    \includegraphics[width=0.86\linewidth]{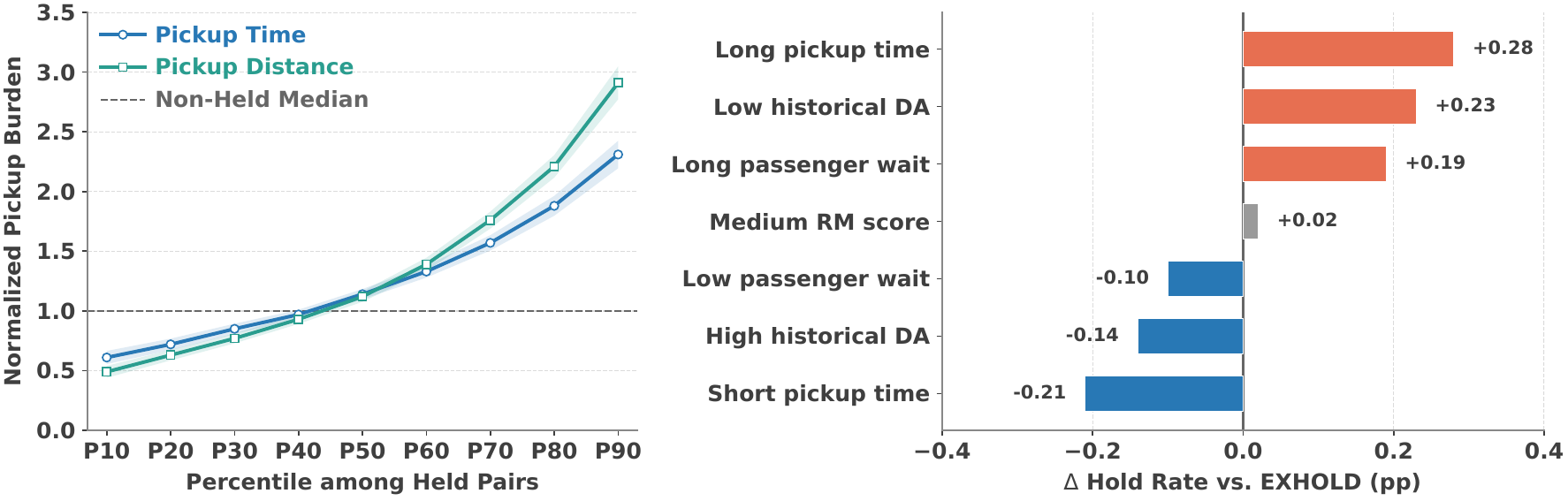}
    \caption{
    Pair-level behavior of \methodname{}.
    \textbf{(a)} Pickup-time and pickup-distance percentiles among held pairs, normalized by the corresponding non-held median.
    \textbf{(b)} Cohort-level hold-rate changes relative to EXHOLD baseline.
    }
    \label{fig:pair_behavior}
\end{figure*}

\subsection{RQ4: Behavioral Analysis}
\label{sec:behavior}

Section~\ref{sec:online_results} shows that \methodname{} improves marketplace outcomes while keeping the aggregate hold ratio nearly unchanged. We therefore examine whether the policy achieves these gains by reallocating interventions across driver--order pairs rather than by increasing holding globally. Figure~\ref{fig:pair_behavior}(a) shows that held pairs tend to have greater pickup burden, particularly in the upper tail: at the $90$th percentile, pickup time and distance reach approximately $2.3\times$ and $2.9\times$ their non-held medians. However, a substantial fraction of held pairs remains below the non-held reference. Therefore, \methodname{} is sensitive to pickup burden without reducing to a fixed pickup-time or distance threshold.

Figure~\ref{fig:pair_behavior}(b) further shows how \methodname{} reallocates holds relative to EXHOLD. It increases holding for long-pickup, long-wait, and low-historical-acceptance cohorts, while reducing holding for their favorable counterparts. The largest shifts occur between the short- and long-pickup cohorts ($-0.21$ and $+0.28$ pp), whereas the medium-score cohort changes little. Together with the stable overall hold ratio, these results indicate that \methodname{} improves the allocation of hold opportunities rather than applying more aggressive intervention globally. Additional temporal and spatial analyses are provided in Appendix~\ref{sec:appendix_behavior}.

\subsection{Application Use and Payoff}
\label{sec:payoff}

Following the large-scale online experiment, \methodname{} has been ramped up and is currently serving production traffic in DiDi's Brazil marketplace. The deployed system applies the learned experience-tier policy to real-time driver--order candidates. Its production operation preserves the improvements observed during experimentation, including higher trip completion, driver income, and acceptance conversion, together with lower passenger cancellations and nearly unchanged overall hold intensity.

The \methodname{} pipeline is designed to keep model development separate from latency-sensitive serving. Multi-view preference construction, RM training, hard-pair sampling, low-identifiability filtering, and simulator-based policy learning are performed offline. Therefore, the RM introduces no additional inference dependency on the online request path. This separation allows the RM, policy, and execution table to be updated and validated independently.

Production monitoring of \methodname{} covers the experience-tier distribution, hold-time schedule, overall hold ratio, and core marketplace indicators, including \CR{}, \DI{}, \DA{}, and cancellation rates. Offline retraining additionally monitors RM-score drift, held-out preference-ranking performance, and the proportion of interactions removed by low-identifiability filtering. Versioned policy artifacts, predefined alarms, and rollback controls protect the marketplace against unexpected distribution shifts or service regressions.
\section{Related Work}
\label{sec:related_work}

\paragraph{Ride-hailing dispatch, matching, and timing control.}

Large-scale ride-hailing platforms require real-time decisions over dynamic passenger--driver markets under tight latency and reliability constraints~\cite{wen2024survey,teusch2023systematic,taylor2024shared}.
Prior work combines learning with planning for dispatch~\cite{zhang2017taxi,xu2018large,ke2021equilibrium,liu2022deep} and models pair-level matching quality~\cite{wang2021secure,suhr2019two}.
Other studies optimize matching intervals or pickup-time targets to balance matching quality, waiting, and cancellation risk~\cite{qin2021optimizing,rong2025satisficing,yan2020dynamic}.
Together, this literature establishes the broader operational context for fine-grained timing control in ride-hailing matching pipelines.

\paragraph{Experience-aware hold control and EXHOLD}
Ride-hailing experience is multi-objective and context dependent, as passenger and driver responses vary with pickup time~\cite{afeche2022ride,feng2021we}, trip attributes, and marketplace conditions~\cite{ashkrof2022ride,wang2021secure}.
Our prior system EXHOLD learns interpretable experience tiers using a contextual bandit and maps them to hold times through guardrail-constrained optimization~\cite{liu2026exhold}.
However, its policy relies on a globally handcrafted reward, which requires designers to specify and tune the trade-offs among completion, cancellations, waiting, and driver effort~\cite{hadfield2017inverse}.
\methodname{} retains EXHOLD's production-tested policy architecture and execution guardrails, but learns context-dependent supervision from implicit passenger--driver preferences and filters weakly identifiable feedback.
It thus advances EXHOLD from \emph{experience-aware reward engineering} to \emph{preference-based experience alignment}.

\paragraph{Pairwise preference learning and reward modeling.}
Pairwise preference learning represents supervision through relative comparisons rather than absolute utility labels, including ranking from implicit feedback~\cite{rendle2012bpr, liu2024interact} and reward learning from trajectory comparisons~\cite{christiano2017deep, liu2025online}.
This paradigm has also been adopted in language-model alignment, where pairwise rankings of candidate responses supervise a reward model that subsequently guides policy optimization~\cite{ouyang2022training,stiennon2020learning}.
In ride-hailing, however, explicit comparisons under the same decision context are unavailable; \methodname{} instead constructs implicit preference pairs from related order, driver, and market interactions.
Because such behavioral comparisons can be noisy and preference learning is sensitive to corrupted labels~\cite{cheng2024rime,ibarz2018reward, yang2026road}, the learned reward model provides preference-based supervision for robust hold-policy learning.

\section{Conclusion}
\label{sec:conclusion}

We presented \methodname{}, a production-scale framework that advances real-time hold control from handcrafted reward engineering to experience alignment learned from implicit marketplace behavior. \methodname{} constructs multi-view preferences, improves RM learning through balanced and model-adaptive sampling, and uses the learned RM to provide dense rewards while filtering low-identifiability feedback during policy learning. Large-scale online experiments and full deployment in DiDi's Brazil marketplace demonstrate that \methodname{} improves passenger--driver experience and marketplace efficiency through more precise allocation of hold decisions, providing a practical approach to aligning large-scale marketplace policies with heterogeneous experience.

\newpage

\bibliographystyle{ACM-Reference-Format}
\bibliography{sample-base}

\newpage

\appendix

\section{Supplementary Experimental Analyses}
\label{sec:appendix_experiments}

\subsection{Behavioral Analysis}
\label{sec:appendix_behavior}

\subsubsection{Temporal Adaptation}

We provide additional analyses of how \methodname{} adapts its hold decisions under different marketplace conditions.
While Section~\ref{sec:behavior} focuses on pair-level decision patterns, here we further study whether the learned policy adjusts its intervention strategy according to temporal dynamics and local supply--demand environments.

Figure~\ref{fig:appendix_time_temporal_behavior} (a) compares the hold-rate distribution over the daily cycle.
Compared with EXHOLD, \methodname{} does not uniformly increase or decrease intervention intensity.
Instead, it reallocates hold decisions toward periods where matching uncertainty is higher.

In particular, \methodname{} applies relatively stronger holding during late-night and early-morning periods, when marketplace activity is sparse and alternative matching opportunities are limited.
During dense daytime periods, the policy becomes less conservative, avoiding unnecessary delays for already promising interactions.

This adaptive behavior suggests that the aligned policy responds not only to static driver--order compatibility but also to the contextual value of waiting under different marketplace conditions.

\subsubsection{Supply--Demand Regime Adaptation}

\begin{figure}[ht]
    \centering
    \includegraphics[width=\linewidth]{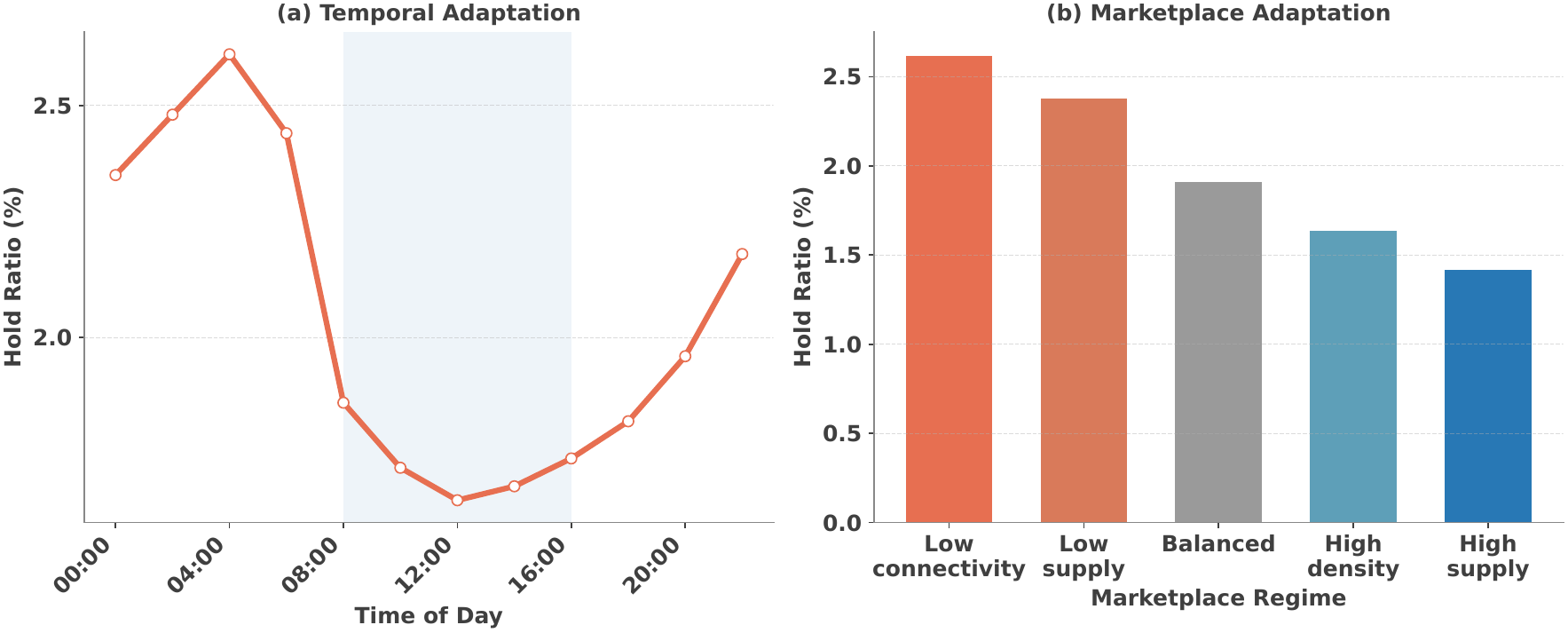}
    \caption{
        Behavioral adaptation of \methodname{} under heterogeneous marketplace conditions.
        (a) The hold ratio varies across time of day.
        (b) The policy adapts to local supply--demand regimes.
        }
    \label{fig:appendix_time_temporal_behavior}
\end{figure}

We further analyze policy behavior under different local supply--demand conditions.
Figure~\ref{fig:appendix_time_temporal_behavior} (b) reports the hold ratio of \methodname{} across marketplace regimes. The policy applies more holding in low-connectivity regions where additional waiting can create meaningful matching opportunities.
As local supply availability and matching density increase, \methodname{} gradually becomes less conservative because immediate matching is more likely to achieve satisfactory outcomes.

This pattern demonstrates that \methodname{} does not optimize the driver--order pair in isolation.
Instead, it considers the opportunity cost of waiting: the value of holding a candidate depends not only on the current interaction quality but also on whether the surrounding marketplace provides alternative opportunities.

Together with the analysis in Section~\ref{sec:behavior}, these results show that \methodname{} learns a context-dependent hold strategy that combines individual experience assessment with marketplace-level adaptation.

\subsection{Reward Model Validation}
\label{sec:rm_validation}

The effectiveness of \methodname{} relies on the quality of the learned reward model, which serves as the alignment signal for downstream policy optimization.
Unlike supervised prediction models, the RM is not designed to estimate an absolute satisfaction probability; instead, it learns a relative ordering over driver--order interactions by leveraging implicit preference supervision extracted from production trajectories.
Therefore, we evaluate the RM from two complementary perspectives:
(i) whether it can recover held-out preference relations, and
(ii) whether its learned score meaningfully correlates with downstream passenger--driver experience outcomes.

\begin{figure}[ht]
    \centering
    \includegraphics[width=\linewidth]{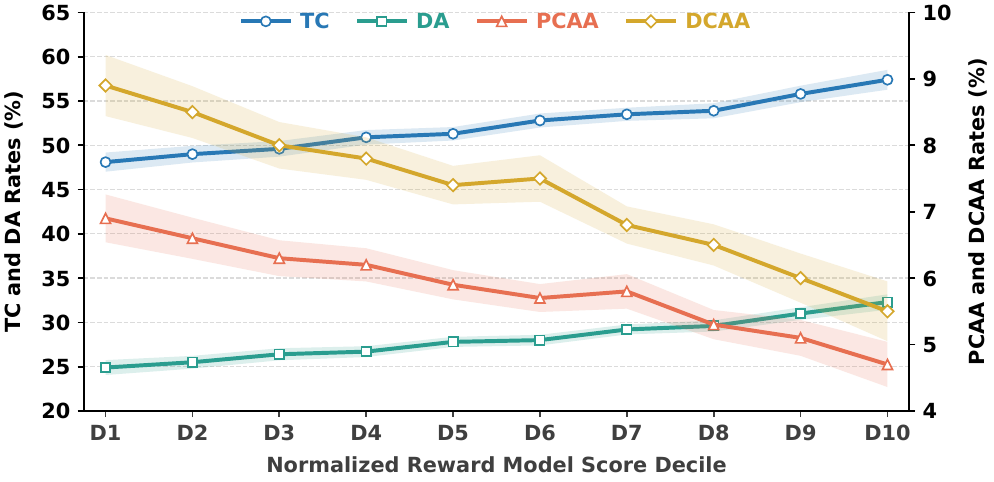}
    \caption{
    Downstream outcomes across normalized RM-score deciles.
    Higher RM scores correspond to higher trip completion and driver
    acceptance rates, while reducing passenger and driver cancellation rates.
    }
    \label{fig:rm_decile_outcomes}
\end{figure}

\subsubsection{Pairwise preference discrimination}

We first evaluate whether the RM preserves the relative ordering encoded in the preference dataset.
Given a held-out preference pair
$\bigl((s^+,a^+),(s^-,a^-)\bigr)$,
the RM is considered correct if
$r_\phi(s^+,a^+)>r_\phi(s^-,a^-)$.
Table~\ref{tab:rm_validation} reports complementary ranking and outcome-discrimination diagnostics on held-out production data. The RM achieves a pairwise preference accuracy of $0.845$, indicating that the learned score consistently recovers the relative experience ordering from production-derived preference supervision.
This result is particularly meaningful since the preference labels are not human annotations but are inferred from naturally occurring marketplace trajectories, which contain substantial behavioral noise and unobserved confounding factors.

\begin{table}[ht]
    \centering
    \caption{
    Reward Model validation results on held-out production data.
    Pairwise accuracy evaluates preference ranking ability, and outcome AUC values are auxiliary diagnostics measuring whether the learned RM score separates favorable and unfavorable downstream outcomes.
    }
    \label{tab:rm_validation}
    \begin{tabular}{lc}
        \toprule
        \textbf{Metric} & \textbf{Value} \\
        \midrule
        Pairwise preference accuracy & 0.845 \\
        Outcome AUC & 0.852 \\
        Completion vs.\ \PCAA{} AUC & 0.864 \\
        Completion vs.\ \DCAA{} AUC & 0.868 \\
        Completion vs.\ driver non-response AUC & 0.815 \\
        \bottomrule
    \end{tabular}
\end{table}

\subsubsection{RM score as an experience ordering signal}

Although the RM is trained only with pairwise preference objectives, its continuous score should ideally reflect a meaningful ordering of marketplace experience.
To examine this property, we normalize RM scores on held-out interactions and partition samples into ten equal-frequency score deciles.

Figure~\ref{fig:rm_decile_outcomes} reports downstream experience outcomes across RM-score deciles.
The results show clear monotonic trends: higher RM-score groups consistently achieve higher trip completion and driver acceptance rates, while experiencing lower passenger and driver cancellation rates.
The aggregate outcome AUC is $0.852$, matching the full multi-view result reported in Table~\ref{tab:view_ablation}; the remaining rows of Table~\ref{tab:rm_validation} disaggregate this diagnostic by representative non-completion outcome.
In particular, the gap between the highest and lowest score deciles is substantial for both positive outcomes and cancellation-related outcomes,
demonstrating that the RM captures fine-grained experience differences rather than merely separating extreme failure cases.

\subsubsection{Discussion on behavioral signal quality}

The discrimination performance varies across different behavioral outcomes. The RM achieves stronger separation for active cancellation events
(\PCAA{} and \DCAA{}) than for driver non-response. This difference is expected because active cancellations provide explicit behavioral feedback, whereas non-response is a weaker signal influenced by multiple operational factors, including driver availability, attention, and external constraints.

Nevertheless, the RM maintains meaningful discrimination even under noisy feedback sources.
This observation supports the design choice of learning from multiple preference views rather than relying on any single behavioral signal.
By aggregating order-centric, driver-centric, and marketplace-context comparisons, \methodname{} converts sparse individual outcomes into a richer relative experience representation that can be used for policy alignment. Overall, these validation results demonstrate that the learned RM provides a stable and interpretable alignment signal.

\section{Implementation Details}
\label{sec:appendix_algorithm}

Algorithm~\ref{alg:align_hold_training} provides a roadmap for the complete training and deployment workflow of \methodname{}. Its major calls correspond to the concrete modules detailed below. \textsc{ConstructMultiViewPairs} creates multi-view preference pair using the rules in Section~\ref{app:preference_rules}. \textsc{TrainRewardModel} fits the preference scorer with model-adaptive sampling. \textsc{CalibrateRewardScale} estimates score normalization statistics on held-out data, \textsc{RolloutSlice} generates simulator interactions under the current policy, and \textsc{DetectSpikeBand} identifies the slice-specific dense-score interval used as a proxy for low-identifiability feedback. Finally, \textsc{EstimateExecutionStats} and \textsc{CalibrateHoldTimes} instantiate the fixed production execution module that maps the learned tiers to guarded hold durations.

At deployment, only the tier policy and the calibrated lookup table are required. Online inference uses neither downstream outcomes nor RM-based filtering: the policy predicts an experience tier, retrieves its hold time in constant time, and passes the result through the existing production execution path.

\begin{algorithm}[H]
\footnotesize
\caption{End-to-End Workflow of \methodname{}}
\label{alg:align_hold_training}
\begin{algorithmic}[1]
\Require Production logs $\mathcal L$, calibration data $\mathcal C$, matching simulator $\mathcal M$, and service constraints $\Gamma$
\Ensure RM $r_\phi$, tier policy $\pi_\theta$, and hold-time table $\boldsymbol{\tau}^{*}$
\State $\mathcal P\gets\Call{ConstructMultiViewPairs}{\mathcal L}$
\State $r_\phi\gets\Call{TrainRewardModel}{\mathcal P,\text{ balanced and adaptive sampling}}$
\State $(\mu_\phi,\sigma_\phi)\gets\Call{CalibrateRewardScale}{r_\phi,\mathcal C}$
\State Initialize tier policy $\pi_\theta$
\For{each simulator training slice $b$}
    \State $\mathcal S_b\gets\Call{RolloutSlice}{\mathcal M,\pi_\theta}$
    \State $\mathcal B_{\mathrm{low}}^{(b)}\gets\Call{DetectSpikeBand}{r_\phi,\mathcal S_b}$
    \For{each simulator interaction $(s_t,a_t)$ in $\mathcal S_b$}
        \State $u_t\gets r_\phi(s_t,a_t)$
        \State $\widetilde r_t\gets(u_t-\mu_\phi)/(\sigma_\phi+\epsilon)$
        \If{$u_t\notin\mathcal B_{\mathrm{low}}^{(b)}$}
            \State $\pi_\theta\gets\Call{UpdateContextualBandit}{\pi_\theta,s_t,a_t,\widetilde r_t}$
        \EndIf
    \EndFor
\EndFor
\State $(F_{i,j},p_{i,j},\mathcal G_i)\gets\Call{EstimateExecutionStats}{\pi_\theta,\mathcal L}$
\State $\boldsymbol{\tau}^{*}\gets\Call{CalibrateHoldTimes}{F,p,\mathcal G,\Gamma}$
\Statex \hspace{\algorithmicindent}\Comment{Quantile discretization and constrained DP with a monotone schedule}
\For{each online decision state $s_t$}
    \State $a_t\gets\pi_\theta(s_t)$
    \State $\tau_t\gets\tau_{a_t}^{*}$
    \State Execute hold duration $\tau_t$
\EndFor
\State \Return $r_\phi$, $\pi_\theta$, and $\boldsymbol{\tau}^{*}$
\end{algorithmic}
\end{algorithm}

\subsection{Multi-View Preference Construction Details}
\label{app:preference_rules}

Here we provide the detailed construction rules for the three preference views introduced in Section~\ref{sec:preference_mining}. The objective is to transform production matching trajectories into relative preference supervision while reducing irrelevant variation between the compared interactions. All three views use the same outcome-based preference semantics but differ in the contextual relationship used to select comparable interactions.

\paragraph{Logged interactions and preference semantics.}
We represent a logged interaction as $e_i=(s_i,a_i,y_i,t_i,o_i,d_i)$, where $s_i$ contains only features available at decision time, $a_i$ is the executed experience tier, $y_i$ is the eventual downstream outcome, $t_i$ is the decision timestamp, and $o_i$ and $d_i$ identify the corresponding order and driver. We use the following mutually exclusive terminal-outcome space:
\begin{equation}
\label{eq:appendix_outcome_space}
\begin{aligned}
    \mathcal Y^-
    &=
    \left\{
        \text{non-}\DA,
        \CBA,
        \PCAA,
        \DCAA,
        \text{other non-}\CR
    \right\}, \\
    \mathcal{Y} &= \{\mathcal{Y}^- + \text{\CR}\}.
\end{aligned}
\end{equation}
Thus, $y_i\in\mathcal Y$, and $\mathcal Y^{-}$ is exactly the non-completion subset used in Equation~\eqref{eq:outcome_partition}. Post-decision information is used only to determine the outcome label $y_i$ and is not included in the RM input. 

We define the completion indicator
\begin{equation}
    q_i=\mathbb I[y_i=\text{\CR}].
\end{equation}
For any pair of contextually related interactions with different completion indicators, the completed interaction is treated as preferred:
\begin{equation}
    q_i=1,\quad q_j=0
    \quad\Longrightarrow\quad
    (s_i,a_i)\succ(s_j,a_j).
\end{equation}
The non-completed interaction may correspond to driver non-response, \CBA, \PCAA, \DCAA, or another non-completion outcome. We do not construct preferences between two completed interactions or between two non-completed interactions. The resulting supervision should be interpreted as an implicit preference for decisions associated with successful joint interaction, rather than an explicit passenger or driver satisfaction label. Contextual pairing is therefore important: it reduces differences unrelated to the driver--order decision before the completion-based label is applied.

\paragraph{Order-centric preference construction.}
For each order $o$, we chronologically organize its broadcast and matching interactions into an order trajectory. Suppose interaction $e_i$ does not complete and a subsequent interaction $e_j$ completes the same order. We construct the preference
\begin{equation}
\label{eq:order_preference}
    (s_j,a_j)\succ(s_i,a_i)
\end{equation}
when
\begin{equation}
    o_i=o_j,\quad
    0<t_j-t_i\leq 10\ \text{minutes},
    \quad
    q_i=0,\quad q_j=1.
\end{equation}

The 10-minute window restricts comparisons to the same local request episode. Within this interval, the origin, destination, and underlying trip intent remain fixed, while the order may encounter different drivers, experience-tier actions, and marketplace states. This view therefore extracts preference evidence from the evolution of a single passenger request. If an order contains multiple eligible non-completed broadcasts before completion, each eligible interaction can contribute a candidate preference pair with the completed interaction. These pairs remain associated with the order-centric pool $\mathcal P_{\mathrm{ord}}$ and are subsequently sampled during RM training rather than assigned additional manually designed weights.

\paragraph{Driver-centric preference construction.}
For each driver $d$, we chronologically organize matching interactions into a driver sequence. A completed and a non-completed interaction form a driver-centric preference pair when
\begin{equation}
\label{eq:driver_preference}
    d_i=d_j,\quad
    |t_i-t_j|\leq 1\ \text{hour},
    \quad
    q_i\neq q_j.
\end{equation}
The interaction with $q=1$ is placed on the preferred side of the pair. The one-hour window is designed to keep comparisons local to the driver's recent operating period. It reduces variation in persistent driver characteristics and limits changes in factors such as working status, local familiarity, and short-term behavioral tendency. At the same time, the compared interactions may involve different orders, pickup burdens, destinations, experience-tier actions, and local supply--demand conditions. The resulting pool $\mathcal P_{\mathrm{drv}}$ therefore provides complementary evidence about how the same driver responds to different matching decisions.

\paragraph{Market-context preference construction.}
The market-context view is constructed from local bipartite matching graphs. For a graph snapshot $G_t=(\mathcal D_t,\mathcal O_t,\mathcal E_t)$, $\mathcal D_t$ and $\mathcal O_t$ denote the active drivers and orders, and $\mathcal E_t$ contains the eligible driver--order edges considered by the matching system. Each edge is associated with its decision-time state, executed experience tier, and eventual outcome.
We compare completed and non-completed interactions from the same local graph snapshot or from immediately adjacent snapshots belonging to the same local dispatch region. Formally, for two eligible edges $e_i$ and $e_j$, we construct
\begin{equation}
\label{eq:market_preference}
    (s_i,a_i)\succ(s_j,a_j)
\end{equation}
when $q_i=1$, $q_j=0$, and their graph snapshots satisfy the production context-matching criteria. Restricting comparisons to the same or adjacent snapshots keeps local supply, demand, and matching opportunities closely aligned. Unlike the order- and driver-centric views, the compared interactions need not share an order or driver. This view therefore increases preference coverage by exploiting contemporaneous alternatives that arise under similar marketplace conditions.

\paragraph{Data partitioning and quality control.}
We partition the production logs chronologically into training, validation, and test periods at a $7{:}2{:}1$ ratio before constructing preference pairs. The validation partition also serves as the held-out calibration set used to estimate $(\mu_\phi,\sigma_\phi)$. Preference pairs are formed only within a single split. Order trajectories are not allowed to cross split boundaries, driver comparisons spanning different splits are discarded, and graph snapshots are assigned according to their decision timestamps. This procedure prevents the same local interaction sequence from appearing in both RM training and evaluation.

We further exclude interactions without a reliable terminal outcome or a complete decision-time state representation. Preference labels are generated only when one interaction completes and the other does not; ambiguous same-class comparisons are discarded. These rules produce three view-specific pools:
\begin{equation}
    \mathcal P_v
    =
    \left\{
        \bigl((s_i^+,a_i^+),(s_i^-,a_i^-)\bigr)
    \right\}_{i=1}^{N_v},
    \qquad
    v\in
    \{\mathrm{ord},\mathrm{drv},\mathrm{mkt}\},
\end{equation}
where the positive and negative superscripts denote the completed and non-completed sides of each preference pair. The pools differ in size because order, driver, and market-context relationships occur at different frequencies in production. To prevent the most frequent source from dominating RM learning, the regular portion of each mini-batch samples the three pools at a $1{:}1{:}1$ ratio. This balanced component accounts for $80\%$ of the mini-batch. The remaining $20\%$ is sampled globally from the $70$th--$90$th percentiles of the current pairwise-loss distribution, as described in Section~\ref{sec:reward_model}. Consequently, the regular component preserves equal multi-view representation, while model-adaptive hard sampling can emphasize informative comparisons regardless of their source view.

\subsection{Reward Model Architecture and Training Settings}
\label{sec:appendix_rm_settings}

This subsection instantiates the \textsc{TrainRewardModel} call in Algorithm~\ref{alg:align_hold_training}. Numerical features are standardized using streaming estimates of their means and variances. Each standardized scalar is then mapped to a 64-dimensional token by a two-layer MLP with a 16-dimensional hidden representation. Categorical features are mapped by stable hashing to $2^{18}$ buckets, embedded in 16 dimensions, and linearly projected to the same 64-dimensional token space. A learned \texttt{[CLS]} token and learned positional embeddings are prepended before Transformer encoding.

The RM uses two pre-norm Transformer layers with four attention heads, hidden size 64, feed-forward size 256, GELU activations, and dropout 0.1. A LayerNorm followed by a $64\!\rightarrow\!64\!\rightarrow\!1$ MLP head produces the scalar preference score. Table~\ref{tab:appendix_rm_settings} summarizes the remaining training settings.

\begin{table}[ht]
    \centering
    \caption{Reward-model architecture and training configuration.}
    \label{tab:appendix_rm_settings}
    \small
    \begin{tabular}{p{0.34\columnwidth}p{0.58\columnwidth}}
        \toprule
        \textbf{Component} & \textbf{Setting} \\
        \midrule
        Numerical tokenizer & MLP $1\!\rightarrow\!16\!\rightarrow\!64$ with GELU \\
        Categorical tokenizer & $2^{18}$ hash buckets; 16-d embedding; 64-d projection \\
        Transformer encoder & 2 layers; 4 heads; hidden size 64; FFN size 256; dropout 0.1 \\
        Score head & LayerNorm and MLP $64\!\rightarrow\!64\!\rightarrow\!1$ \\
        Optimizer & AdamW; learning rate $3\times10^{-4}$; weight decay $10^{-4}$ \\
        Stabilization & Mixed precision; gradient clipping at 1.0; seed 42 \\
        Pairwise objective & Bradley--Terry loss in Equation~\eqref{eq:rm_loss} with unit temperature \\
        Regular sampling & 80\% of each mini-batch; view ratio $1{:}1{:}1$ \\
        Adaptive hard sampling & 20\% of each mini-batch; global $70$th--$90$th pairwise-loss percentiles \\
        Data separation & Chronological $7{:}2{:}1$ training, validation/calibration, and test partitions; no cross-split pair \\
        \bottomrule
    \end{tabular}
\end{table}

For the regular component, each mini-batch draws equally from $\mathcal P_{\mathrm{ord}}$, $\mathcal P_{\mathrm{drv}}$, and $\mathcal P_{\mathrm{mkt}}$. For the adaptive component, we recompute Equation~\eqref{eq:rm_loss} under the current RM, form the loss distribution over the mixed preference pool, and sample pairs from its $70$th--$90$th percentiles. Selection is therefore based on current pairwise loss rather than an outcome-specific score threshold. The hard component supplies 20\% of each mini-batch and the balanced regular component supplies the remaining 80\%. We optimize only the pairwise Bradley--Terry objective and disable the auxiliary pointwise binary-classification loss.

\subsection{Low-Identifiability Filtering and Tier-Policy Settings}
\label{sec:appendix_policy_settings}

This subsection instantiates \textsc{CalibrateRewardScale}, \textsc{DetectSpikeBand}, and the simulator-interaction loop in Algorithm~\ref{alg:align_hold_training}. After RM training, we freeze $r_\phi$ and evaluate it in mini-batches of 4,096. On the held-out calibration partition, we compute $\mu_\phi$ and $\sigma_\phi$ and normalize every score according to Equation~\eqref{eq:normalized_reward}. The normalized score itself is the contextual-bandit reward; no handcrafted outcome reward or tier-dependent action weight is added.

\paragraph{Dynamic score-spike identification.}
For each policy-training slice, the frozen RM first scores all interactions, after which the rejection band is estimated from the training portion only. Let $u_i=r_\phi(s_i,a_i)$ be the finite raw scores. We form a 256-bin equal-width histogram over $[Q_{0.005}(u),Q_{0.995}(u)]$ and take the most populated bin $k^*$ as the candidate peak. Starting from $k^*$, we extend to the maximal contiguous bin range $[\ell,r]$ whose counts remain at least half of the peak count. With histogram edges $e_j$, the rejection band is
\begin{equation}
\begin{aligned}
    c
    &=
    \frac{e_{k^*}+e_{k^*+1}}{2},\\
    w
    &=
    \operatorname{clip}\!\left(
        \frac{e_{r+1}-e_\ell}{2},\,0.005,\,0.05
    \right),\\
    \mathcal B_{\mathrm{low}}^{(b)}
    &=
    [c-w,c+w].
\end{aligned}
\label{eq:appendix_spike_band}
\end{equation}
The band is activated only when the peak count is at least three times the median count of nonempty bins and the band contains at least $0.2\%$ of the finite scores. Slices with fewer than 128 finite scores are left unfiltered; equivalently, $\mathcal B_{\mathrm{low}}^{(b)}=\varnothing$ when the activation criteria are not met. These thresholds are fixed across experiments. The detector is label-free and re-estimates the score location for every training slice; samples in the detected band are excluded from policy updates, while the held-out portion is not filtered.

During interaction with the matching simulator, the policy chooses $a_t=\pi_\theta(s_t)$ and receives $\widetilde r_\phi(s_t,a_t)$. The interaction is retained exactly when the indicator $m_t$ in Equation~\eqref{eq:retention} equals one. Retained interactions update the contextual bandit with the normalized RM score, whereas interactions whose raw RM scores fall in $\mathcal B_{\mathrm{low}}^{(b)}$ do not update either the encoder or the bandit head. The RM score is not appended to the policy context, and neither RM inference nor low-identifiability filtering appears on the online serving path.

The retained data train a four-action Transformer--LinUCB policy: immediate release and three increasing hold tiers. The context vector is projected to 192 dimensions, encoded by two Transformer layers with six attention heads and dropout 0.1, and mapped to a 64-dimensional $\ell_2$-normalized representation. Table~\ref{tab:appendix_policy_settings} reports the policy-training configuration.

\begin{table}[ht]
    \centering
    \caption{Experience-tier policy and filtering configuration.}
    \label{tab:appendix_policy_settings}
    \small
    \begin{tabular}{p{0.34\columnwidth}p{0.58\columnwidth}}
        \toprule
        \textbf{Component} & \textbf{Setting} \\
        \midrule
        Action space & 4 tiers: immediate release and 3 hold levels \\
        Policy encoder & 2 Transformer layers; 6 heads; hidden size 192; dropout 0.1 \\
        Output representation & 64 dimensions with $\ell_2$ normalization \\
        LinUCB parameters & UCB exploration coefficient $\beta_{\mathrm{UCB}}=1$; ridge coefficient $\lambda=1$ \\
        Encoder optimizer & Adam; learning rate $7\times10^{-4}$; gradient clipping at 1.0 \\
        Replay and refresh & Buffer size 200,000; refresh every 10,000 updates; 200 steps per refresh; batch size 512 \\
        Exploration warm-up & 100,000 random-action updates \\
        Policy training & 100 epochs; 10,000 simulator interactions per training slice \\
        RM reward & Held-out mean--standard-deviation normalization in Equation~\eqref{eq:normalized_reward} \\
        Low-ID detection & Per-slice, label-free score histogram; 256 bins over the 0.5th--99.5th percentiles \\
        Rejection criterion & Contiguous half-peak band; prominence $\geq3$; score mass $\geq0.2\%$ \\
        Update rule & Retain $m_t=1$ interactions and update with $\widetilde r_\phi(s_t,a_t)$ \\
        Serving path & Tier policy and hold-time lookup only; no RM or filtering \\
        \bottomrule
    \end{tabular}
\end{table}

Once simulator training is complete, the policy produces one of four ordered experience tiers. The remaining two offline calls in Algorithm~\ref{alg:align_hold_training} translate these discrete decisions into executable durations. This final calibration stage is intentionally separated from representation and policy learning: it reuses the production-tested EXHOLD module and can be refreshed, audited, or rolled back without retraining either learned model.

\subsection{Guardrail-Constrained Hold-Time Calibration}
\label{sec:appendix_hold_calibration}

\methodname{} changes how experience tiers are learned, but not how those tiers are translated into executable hold durations. We reuse, without modification, the production hold-time calibration and execution module introduced in EXHOLD~\cite{liu2026exhold}. Keeping this module fixed preserves the service guardrails and operational properties already validated in production, while isolating the contribution of the learned experience-alignment policy.

Given a fixed tier policy $\pi_\theta$, we route recent production samples to tiers $i\in\{0,\ldots,K\}$. Let $T$ denote the system-defined execution clock and $y\in\mathcal Y$ the observed downstream outcome. For each tier--outcome pair $(i,j)$, we estimate the empirical cumulative distribution function and routing fraction
\begin{equation}
    F_{i,j}(u)=\Pr(T\le u\mid a=i,y=j),
    \qquad
    p_{i,j}=\frac{N_{i,j}}{\sum_{i'=0}^{K}N_{i',j}},
    \label{eq:appendix_execution_stats}
\end{equation}
where $N_{i,j}$ is the number of routed samples with tier $i$ and outcome $j$. These statistics can be refreshed from recent logs without retraining the policy.

We then choose one hold duration $\tau_i$ for each tier by solving the same guardrail-constrained problem as EXHOLD:
\begin{equation}
\label{eq:appendix_hold_objective}
\begin{aligned}
    \max_{\{\tau_i\}}\quad
    &\sum_{i=0}^{K}\sum_{j\in\mathcal Y}
        p_{i,j}v_jF_{i,j}(\tau_i)\\
    \text{s.t.}\quad
    &\sum_{i=0}^{K}p_{i,\text{\CR}}F_{i,\text{\CR}}(\tau_i)
        \le\delta_{\mathrm{TC}},\\
    &0\le\tau_0\le\tau_1\le\cdots\le\tau_K\le\tau_{\max}.
\end{aligned}
\end{equation}
where $v_j$ is the outcome utility weight inherited from EXHOLD, and $\delta_{\mathrm{TC}}$ is the trip-completion guardrail that limits unnecessary holding of interactions that would lead to trip completion. The monotonicity constraint ensures that tiers indicating a stronger need to defer a pair are never assigned shorter hold durations.

Because empirical CDFs are step functions and the number of tiers is small, each $\tau_i$ is restricted to a finite quantile grid
\begin{equation}
    \mathcal G_i=\{Q_i(q)\mid q\in\mathcal Q\},
    \label{eq:appendix_quantile_grid}
\end{equation}
where $Q_i(q)$ is the empirical $q$-quantile of $T$ among tier-$i$ samples and $\mathcal Q$ is a fixed set of quantile levels. The discretized problem is solved efficiently by dynamic programming under the monotonicity and guardrail constraints, yielding an auditable lookup table $\boldsymbol{\tau}^{*}$ that can be versioned and rolled back independently.

At serving time, the learned policy predicts a tier $a_t$, the system retrieves $\tau^{*}_{a_t}$ in constant time, and existing runtime guardrails produce the final executable duration. Neither reward-model inference nor low-identifiability filtering is required online.

\subsection{Simulator Fidelity Evaluation}
\label{app:simulator_fidelity}

We evaluate simulator fidelity by comparing the behavior of the deployed EXHOLD policy under simulation with its observed behavior in production over a held-out seven-day period. The simulator reproduces the overall hold ratio with a small discrepancy, while the mean absolute difference in experience-tier shares is $0.46$ percentage points. Across city--hour cohorts, simulated and online hold rates exhibit a Pearson correlation of $0.94$, and the simulator preserves the same behavioral ordering across major operating regimes: EXHOLD applies more holding to long-pickup and long-wait pairs and less holding under dense matching conditions. The relative prevalence of major downstream outcomes among held interactions is also consistent. These results indicate that the simulator captures the principal state distribution and policy-response patterns required for comparative policy training.

\end{document}